\documentclass[authoryear,preprint,review,12pt]{elsarticle}

\usepackage{amssymb}
\usepackage{amsmath}
\usepackage{amsthm}
\usepackage{amsfonts}
\usepackage{algorithmic}
\usepackage{algorithm}
\usepackage{array}
\usepackage{textcomp}
\usepackage{stfloats}
\usepackage{url}
\usepackage{multirow}
\usepackage{verbatim}
\usepackage{graphicx}
\usepackage{cite}
\usepackage{siunitx}
\usepackage{booktabs}
\usepackage{threeparttable}
\usepackage{subcaption}
\usepackage{placeins}
\usepackage{eurosym}

\usepackage{xcolor}

\journal{Robotics}

\begin{document}

\begin{frontmatter}

\title{Boosting Deep Reinforcement Learning with Semantic Knowledge for Robotic Manipulators}

\author[1]{Luc\'{i}a G\"{u}itta-L\'{o}pez \corref{cor1}}
\ead{lucia.guitta@iit.comillas.edu}

\author[2]{Vincenzo Suriani}
\ead{vincenzo.suriani@unibas.ii}

\author[1]{Jaime Boal}
\ead{jaime.boal@iit.comillas.edu}

\author[1]{\'{A}lvaro J. L\'{o}pez-L\'{o}pez}
\ead{alvaro.lopez@iit.comillas.edu}

\author[3]{Daniele Nardi}
\ead{nardi@diag.uniroma1.it}

\cortext[cor1]{Corresponding author}

\address[1]{Institute for Research in Technology~(IIT), ICAI School of Engineering, Comillas Pontifical University,
			Rey Francisco, 4,
			28008,
			Madrid,
			Spain}
\address[2]{University of Basilicata, via dell'Ateneo Lucano, 10, 85100, Potenza, Italy}
\address[3]{Sapienza University of Rome, Department of Computer, Control and Management Engineering, via Ariosto, 25, 00185, Rome, Italy}

%% Abstract
\begin{abstract}
Deep Reinforcement Learning (DRL) is a powerful framework for solving complex sequential decision-making problems, particularly in robotic control. However, its practical deployment is often hindered by the substantial amount of experience required for learning, which results in high computational and time costs. In this work, we propose a novel integration of DRL with semantic knowledge in the form of Knowledge Graph Embeddings (KGEs), aiming to enhance learning efficiency by providing contextual information to the agent. Our architecture combines KGEs with visual observations, enabling the agent to exploit environmental knowledge during training. Experimental validation with robotic manipulators in environments featuring both fixed and randomized target attributes demonstrates that our method achieves up to {60}{\%} reduction in learning time and improves task accuracy by approximately 15 percentage points, without increasing training time or computational complexity. These results highlight the potential of semantic knowledge to reduce sample complexity and improve the effectiveness of DRL in robotic applications.

This article was accepted and published in Robotics. Cite as: G\"{u}itta-L\'{o}pez, L.; Suriani, V.; Boal, J.; L\'{o}pez-L\'{o}pez, Á.J.; Nardi, D. Boosting Deep Reinforcement Learning with Semantic Knowledge for Robotic Manipulators. Robotics 2025, 14, 86.  \url{https://doi.org/10.3390/robotics14070086}
\end{abstract}

%% Keywords

\end{frontmatter}

%\linenumbers

%% main text
%%

\section{Introduction}
Deep Reinforcement Learning (DRL)~\citep{Lazaridis2020} ~\citep{Lavet2018} is now consolidating as one of the most promising solutions for solving complex sequential decision problems where an agent interacts with its environment. While previous Reinforcement Learning (RL)~\citep{Sutton2018} is limited by its memory and computational capabilities, DRL overcomes these limitations by leveraging the potential of deep learning for function approximation and representation learning in high-dimensional state spaces~\citep{Goodfellow2016}~\citep{Lecun2015}.

The application of DRL to tasks performed by robotic manipulators has broadened the scope from mostly repetitive jobs in fixed scenarios to tasks in changing environments with potentially unknown dynamics~\citep{Panzer2022}~\citep{Rupprecht2022}~\citep{Singh2022}~\citep{Qureshi2018}. One of the major disadvantages of DRL is the extensive amount of experience, i.e., interactions between the agent and the environment, required for learning. It should also be noted that the greater the complexity of the environment, the more experience is needed. Virtual environments can be used to reduce the time and economic cost of overruns. By modeling assets in virtual scenarios, the \textit{sample efficiency} problem is mitigated, as experience is acquired more quickly in simulators. However, the issue of the sheer amount of experience required remains.

\begin{figure}[h!]
\centering
\includegraphics[width=0.5\columnwidth]{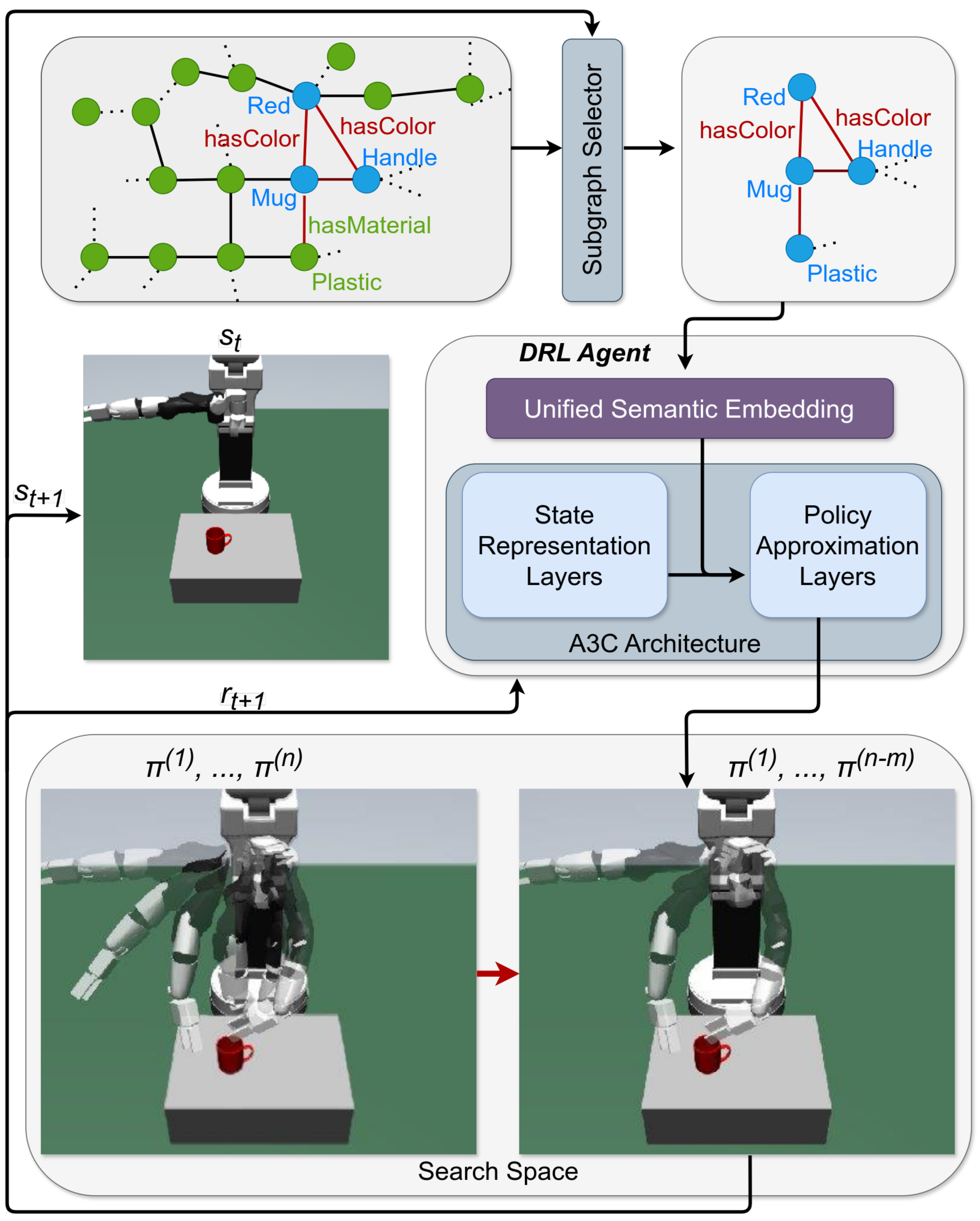}
\caption{Proposed framework diagram. A subgraph selector is used to extract the important entities and relationships based on the environment from a complete graph. This subgraph is embedded and concatenated in the layer prior to the policy approximation blocks of the DRL agent architecture. The result is an improvement in learning times and accuracy and a reduction of the required exploration.}
\label{fig:proposal}
\end{figure}

As Figure~\ref{fig:proposal} presents, we propose integrating contextual information about the environment within the agent's learning architecture to address the sample efficiency issue and enhance the agent's performance. Our main motivation is that any learning process, especially in DRL setups, should improve if the agent has knowledge about the environment that complements its observations. As already shown in~\citep{Davidson2020}, providing some basic semantic structure to the agent can speed up its learning. Ultimately, we argue that this approach introduces \textit{commonsense} as semantic knowledge without additional complexity cost, with only a slight increase in the computational time. The environment's semantic information is obtained from a pre-built knowledge graph, where we extract the embeddings. After the representation learning layers, these embeddings are concatenated as inputs in the agent's architecture. To sum up, our main contributions and findings are:
\begin{itemize}
    \item An original DRL agent architecture that successfully integrates Knowledge Graph Embeddings (KGEs) during the learning process. We modify the baseline layers to allow for concatenating the KGEs with the hidden activations from the ``visual layers''.
    \item A methodology to quantify the improvement of the DRL agents' performance, in terms of time and accuracy, through the use of KGEs with respect to a baseline agent with no semantic information of the environment and an agent with only partial information.
    \item An analysis of the possible improvement of the agents' exploration and exploitation capabilities by examining the distribution of each joint orientation during the evaluation episodes. 
    \item A quantitative and qualitative evaluation of the embedding's influence in different environments and with various robot manipulators.
\end{itemize}

The paper is structured as follows: Section~\ref{sec:RW} presents the related work, Section~\ref{sec:EnvironmentSetup} defines the experiments done, the setup employed, and the MDP designed. In Section~\ref{sec:Methodology}, we describe the methodology followed for the knowledge graph and embedding obtention, as well as the deep reinforcement learning framework, that is, the algorithm used and our architecture, besides the corresponding training and post-training evaluation procedures. Section~\ref{sec:Results} gathers the results discussed in Section~\ref{sec:Discussion}. 
Section~\ref{sec:Limitations} focuses on describing shortcomings and open issues related to our approach, and finally, Section~\ref{sec:Conclusion} sums up the most relevant insights obtained with this research.

%%%%%%%%%%%%%%%%%%%%%%%%%%%%%%%%%%%%%%%%%%
\section{Related Work}
\label{sec:RW}
To leverage the capability of robots to interact with the physical world, robot learning has emerged as a major research area~\citep{kroemer2021review}. Progress in deep learning has contributed significantly to advancements in robot learning, particularly when state representation includes images~\citep{cabi2019scaling}. A variety of algorithms have been developed for robot learning, with reinforcement learning and imitation learning remaining the dominant approaches~\citep{wang2022equivariant}~\citep{zeng2022robotic}. To facilitate the comparison of different reinforcement learning algorithms, numerous benchmarks have been introduced. For instance, some benchmarks focus on specific tasks such as door opening, furniture assembly, and in-hand dexterous manipulation~\citep{lee2021ikea}. Others, like those in~\citep{zhu2020robosuite} and~\citep{delhaisse2020pyrobolearn}, offer diverse environments but lack long-horizon tasks and lack of background knowledge. Background knowledge has been successfully deployed when dealing with natural language action space, to reduce the search space (\citep{ammanabrolu2020graph}~\citep{dambekodi2020playing}), but not extensively applied in the field of robotics.

%Among the most comprehensive benchmarks we have RLBench \citep{james2020rlbench} and BulletArm \citep{wang2022bulletarm}, but when using benchmarking environments, often it is hard to collect

Recently, commonsense has been integrated into the RL pipelines to enrich the representation of the environments. With the advent of the LLMs, a taxonomy of the combination of the two fields of study has been proposed in~\citep{pternea2024rl}, where three classes are defined based on the way that the two model types interact with each other. An LLM can be used to supplement the training of an RL model that performs a general task that is not inherently related to natural language, as \emph{LLM4RL} states.

Here, the semantic knowledge of Large Language Models (LLMs) is leveraged to enhance the performance of Reinforcement Learning (RL) agents ~\citep{Wang2025}. This enhancement can occur in two primary ways: by grounding the agent's environment to improve its performance \citep{xie2023text2reward}~\citep{carta2023grounding}~\citep{Ma2024}, or by improving the training process of the RL agent. Training an RL agent is often computationally intensive and requires significant resources and large amounts of data. Moreover, RL training can suffer from inefficient sampling, particularly for complex, long-term tasks. Consequently, several LLM4RL frameworks aim to enhance training efficiency, ensuring the successful execution of target tasks during testing. They achieve this by facilitating exploration \citep{quartey2023exploiting}~\citep{du2023guiding}, enabling policy transfer of trained models \citep{reid2022can}, and implementing effective planning to reduce data requirements.

The role of a large-scale model in helping an RL agent, is discussed in \citep{dasgupta2023collaborating}, where the authors investigate how to combine these complementary abilities in a single system consisting of three parts: a Planner, an Actor, and a Reporter. %The Planner is a pre-trained language model that can issue commands to a simple embodied agent (the Actor), while the Reporter communicates with the Planner to inform its next command. 

In the learning process of a grasping task, it has been demonstrated how
generating natural human grasps needs to consider not only the object geometry, but also semantic information. In fact, in \citep{li2024semgrasp}, a semantic-based grasp generation method, termed SemGrasp, generates a static human grasp pose by incorporating semantic information into the grasp representation. 

If LLMs offer an invaluable opportunity to provide semantic information, it is also true that they can be demanding in many applications. To this end, we adopted a lighter architecture, using graphs. Semantic knowledge can successfully be stored in graph representations and manipulated using knowledge graph embedding techniques and pre-trained models.    
%
%The extraction of the information from graphs, i.e., the graph mining tasks, arise from many different application domains, including social networks, biological networks, transportation, and E-commerce, which have been receiving great attention from the theoretical and algorithmic design communities in recent years, and 
Some pioneering work has employed research-rich reinforcement learning (RL) techniques to address graph mining tasks. 
In \citep{nie2023reinforcement}, a unified graph reinforcement learning (GRL) formulation is proposed. However, most existing formulations struggle to effectively integrate groups of semantically related nodes within the RL paradigm. %The pure representation of the  TODO VIncenzo from here
On the other hand, far from the RL field, a semantic representation framework based on a knowledge graph has been demonstrated to be suitable as a method to extract manipulation knowledge from multiple sources of information \citep{miao2023semantic} and guide the agent to generate semantic representations of entities and relations in the knowledge base. 

We propose to leverage the semantic knowledge of a lightweight graph representation to improve sample efficiency, and agent performance in terms of time and accuracy. Our architecture provides additional contextual information about the environment, which complements the agent’s observations, leading to faster learning and higher accuracy,
%. The experiments conducted demonstrate that agents equipped with graph commonsense not only learn faster but also achieve better performance in terms of task success rates 
compared to baseline models without knowledge graphs.

%%%%%%%%%%%%%%%%%%%%%%%%%%%%%%%%%%%%%%%%%%
\section{Environment Setup}
\label{sec:EnvironmentSetup}
\subsection{Experiment description}
The task addressed involves using a robotic arm to approach three targets (e.g., a mug, a bottle, and a cereal box) that move randomly in the workspace area between episodes. The robot's end-effector should reach the grasping point within a distance of \SI{5}{cm} during training and \SI{10}{cm} during the post-training evaluation, with a specific orientation. The training orientation threshold is $\pm15^o$, whereas in the post-training evaluation, it is $\pm20^o$. The placement of the reachable point and the gripper approach orientation depend on the target. The environment observation is a 64$\times$64$\times$3 RGB image without additional proprioception information. We performed the experiments in simulation to enable a thorough analysis of the proposed approach and to allow for total control of all the external drivers that might affect the agent's performance. In this way, we ensure total control of all the external drivers that might decrease the agents' performance.

\subsection{Setup description}
\begin{figure}[h!]
\centering
\includegraphics[width=0.5\columnwidth]{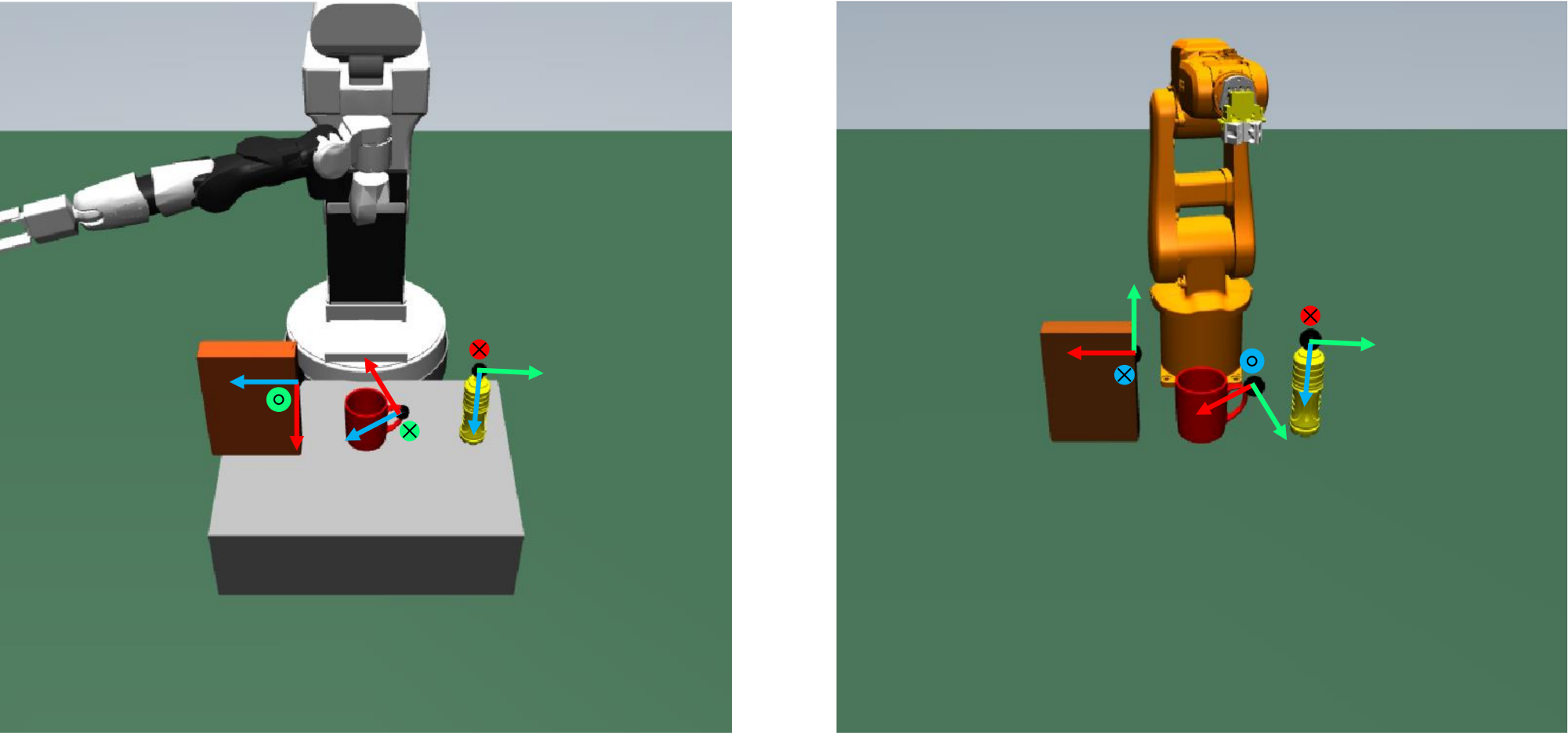}
\caption{Environment observation with the three targets for the TIAGo and IRB120 scenarios. Note that we have added a black sphere on each grasping point for convenience. During training, the resolution is 64$\times$64, and the sphere does not appear. The site axes correspond to the $x$ coordinate in red, the $y$ coordinate in green, and the $z$ coordinate in blue.}
\label{fig:envObsvTargetAxes}
\end{figure}

Two robots have been used in this research. On the one hand, the TIAGo from PAL Robotics~\citep{Pages2016}, a 7-DoF mobile manipulator equipped with a two-finger gripper and mainly designed for human-robot interaction. On the other hand, the IRB120~\citep{IRB120} from ABB, a 6-DoF industrial robotic arm also with a two-finger gripper end-effector. The virtualization of both assets has been carried out through the MuJoCo~\citep{Todorov2012} platform. The arena is the same for both robots. Due to the morphology of the TIAGo arm and to facilitate the reach, it was necessary to add a table that allowed us to place the objects at a greater height in the robot's workspace. Figure~\ref{fig:envObsvTargetAxes} displays the virtual environment for the TIAGo and IRB120 models.

The targets are represented by three graspable objects: a mug, a bottle, and a cereal box. Figure~\ref{fig:envObsvTargetAxes} shows the reachable spots for the three targets. We present the grasping point as a black sphere for better understanding of the reader, but in the observation received by the agent, it is not shown. Besides, Figure~\ref{fig:envObsvTargetAxes} illustrates the orientations of the grasping point axes that the grippers must align with, where the $x$ coordinate is red, the $y$ coordinate is green, and the $z$ coordinate is blue. In the case of the mug, the grasping point is placed on its handle within an inclined orientation with respect to the base plane. For the bottle, we placed the grasping point on the stopper perpendicular to the base plane. Lastly, in the cereal box, the grasping point is set on the right-hand side edge, parallel to the base plane, above half its height. For the experiments without domain randomization (DR) applied to the targets' color, the mug is red (RGB code (1.0, 0.0, 0.0)), the bottle is yellow (RGB code (1.0, 1.0, 0.0)), and the cereal box is brown (RGB code (0.55, 0.27, 0.07)). When we experiment with DR applied to the targets' color, the mug can be red or blue (RGB code (0.0, 0.0, 1.0)), the bottle yellow or purple (RGB code (0.4, 0.0, 0.9)) and the cereal box brown or light-blue (RGB code (0.5, 0.7, 0.9)).

\subsection{MDP definition}
DRL problems can be defined as Markov Decision Processes (MDPs), which are characterized by a set of finite states, actions, a state transition probability matrix, a reward function, and the discount factor. Table~\ref{tab:MDP_def} gathers the MDP definition of our problem. The state is a 64$\times$64$\times$3 RGB image captured at the beginning of each step. We use an episodic setup where each episode lasts 50 steps or less if the robot's gripper reaches the target within the rewarding distance and orientation. The reward given to the agent depends on the relative distance and orientation between the gripper and the grasping point. If the target is not reached, the agent receives a negative reward, whereas if it is, it receives a high positive value, as shown in Table~\ref{tab:MDP_def}, and the episode ends. The action set is formed by seven possible actions that scale the maximum joint movement range (Maximum Position Increment - MPI). At the beginning of the episode, the first and second joint orientations are set randomly within $\pm15\%$ of their working range. Additionally, before starting the episode, one of the three targets is selected and placed randomly following a uniform distribution within the workspace area, which is a 20$\times$\SI{25}{cm} rectangle for the TIAGo environment and a 60$\times$\SI{20}{cm} rectangle for the IRB120 setup.

\begin{table}[h!]
\centering
\renewcommand{\arraystretch}{1.5}
\caption{MDP definition. $\text{rel}_{\text{dist}}$ and $\text{rel}_{\text{deg}}$ represent the relative distance and degrees between the gripper and the grasping point. MPI is the Maximum Position Increment. WRLL and WRUL are the Working Range Lower Limit and the Working Range Upper Limit of the robots. The target position is selected according to a uniform distribution.}
\resizebox{0.5\columnwidth}{!}{
\begin{tabular}{l|l}
    \multicolumn{2}{c}{\textbf{MDP Definition}} \\ \midrule\midrule
    Ep. Length & 50 steps maximum \\ \hline
    \multirow{2}{*}{Reward configuration} & $-2*\text{rel}_{\text{dist}}^{2}-\text{rel}_{\text{deg}}/70$ \\
                                          & 100 if grasping point reached \\ \hline
    \multirow{2}{*}{Reward constraints} & $\text{rel}_{\text{dist}}$ $<$ \SI{5}{cm} \\ 
                                        & $\text{rel}_{\text{deg}}$ $<$ $15^o$ \\ \hline
    \multirow{4}{*}{Action set} & $0$ \\
                                & $\pm \text{MPI}$ \\ 
                                & $\pm \text{MPI}/10$ \\ 
                                & $\pm \text{MPI}/100$ \\ \hline
    \multirow{2}{*}{Initial robot config} & $U(15\% \text{WRLL}, -15\% \text{WRUL})$ \\ 
                                       & for joints 1 and 2 \\ \hline
    \multirow{2}{*}{Initial target config} & x $\sim U(x_{min}, x_{max})$ \\ 
                                       & y $\sim U(y_{min}, y_{max})$ \\ \hline
\end{tabular}}
\label{tab:MDP_def}
\end{table}

%%%%%%%%%%%%%%%%%%%%%%%%%%%%%%%%%%%%%%%%%%
\section{Methodology}
\label{sec:Methodology}
\subsection{Knowledge Graph and Embeddings}
To provide external knowledge to the learner, we extract contextual information of the scene from a Knowledge Graph $\mathcal{G}$. Then, we embedded the information in a latent space in the form of Knowledge Graph Embedding (KGE).

The objective of the KGE is to learn a continuous vector representation of a Knowledge Graph $\mathcal{G}$, which encodes vertices that represent entities $\mathcal{E}$ as a set of vectors $v_{\mathcal{E}} \in \mathbb{R}^{|\mathcal{E}| \times d_{\mathcal{E}}}$ (where $d_{\mathcal{E}}$ is the dimension of the vector of entities $\mathcal{E}$), and as a set of edges which represent relations $\mathcal{R}$ as mappings between vectors $W_{\mathcal{R}} \in \mathbb{R}^{|\mathcal{R}| \times d_{\mathcal{R}}}$, where  $d_{\mathcal{R}}$ is the dimension of the vector of relations. The knowledge graph $\mathcal{G}$ is composed by triples $(h,r,t)$, where $h,t \in \mathcal{E}$ are the head and tail of the relations, while $r \in \mathcal{R}$ is the relation itself. One example of such a triple is (\emph{mug, hasColor, yellow}). %In literature, there are numerous ways of embedding knowledge in a knowledge graph: \textit{transitional models, rotational models, gaussian models}, and many others. However, independently on what is the class of methods that are used, the embedding is learned by minimizing the loss $\mathcal{L}$ computed on a scoring function $f(h,r,t)$ over the set of triples in the knowledge graph, and over the set of negative triples that are generated by negative sampling over the same graph. 

%GloVe is a popular unsupervised learning algorithm that generates word embeddings by aggregating global word-word co-occurrence statistics from a given corpus. 

We initially used ANALOGY as a direct embedding model.
Since we need to have a semantic representation of the scene of the agent, we defined a unified semantic embedding that is extracted by introducing an operator $\Gamma$ that, given a graph $\mathcal{G'}$, select the subgraph $\mathcal{S}$ from the larger knowledge graph $\mathcal{G}$ ($\mathcal{S} \subseteq \mathcal{G}$). 
This subgraph $\mathcal{S}$ contains all the nodes representing the objects that the agent is perceiving and the node at distance 1 from them. resulting in all relevant entities and their relationships pertinent to the specific scene the agent is interacting with. For instance, if the scene includes a mug, a bottle, and a cereal box, the subgraph $\mathcal{S}$ will contain nodes representing these objects and edges denoting their relationships with other nodes (e.g., (\textit{mug, hasColor, red}), \textit{mug, isConnectedTo, handle}), (\textit{bottle, hasShape, cylindrical}), (\textit{cereal box, isMadeOf, cardboard})).
From the subgraph, a textual representation is obtained by concatenating the labels of the nodes and the edges in a single sentence. The sentence is then converted into a single embedding representing the scene. To effectively, capture the semantic nuances of the scene, we replaced ANALOGY with a pre-trained embedding model, GloVe\citep{pennington2014glove}. With GloVe, we generated a single embedding for the scene.
The use of a pre-train model on a large corpus, ensures that the resulting vectors capture a wide range of semantic nuances. The dimension of these embeddings is set to $10 \times 4$ per word, allowing for a rich representation of the subgraph's content. 
%This embedding process effectively captures the semantic meanings and relationships among the entities within the subgraph, resulting in a set of Knowledge Graph Embeddings (KGEs).

The resulting GloVe-based embeddings are then integrated into the learner's architecture. In our proposed setup, the embeddings are concatenated with the hidden activations from the visual observation layers before the policy approximation block. This strategic placement allows the DRL agent to utilize both the raw visual data and the rich contextual information provided by the embeddings, enhancing its ability to understand and interact with the environment.

\subsection{Deep Reinforcement Learning Framework}
\subsubsection{Asynchronous Advantage Actor Critic (A3C)}
For the sake of clarity, we first recap the concepts and terminology used to describe DRL. Model-free DRL includes value-based and policy-based approaches. The former are methods that learn either the estimation of the expected discounted cumulative reward, represented by the state-value function, $V^{\pi}(s)$, or the estimation of the expected discounted cumulative reward for taking action $a$ in state $s$ at step $t$, defined with the action-state value function, $Q^{\pi}(s, a)$ whose definitions are:
\begin{equation}
V^{\pi}(s) = \mathbb{E}_{\pi} \left[ \sum_{t=0}^{\infty} \gamma^t r_{t+1} \mid s_t = s \right]
\end{equation}
\begin{equation}
Q^{\pi}(s, a) = \mathbb{E}_{\pi} \left[ \sum_{t=0}^{\infty} \gamma^t r_{t+1} \mid s_t = s, a_t = a \right]
\end{equation}
where $\mathbb{E}_{\pi}$ is the expectation following policy $\pi$, $\gamma$ is the discount factor, $r_{t+1}$ is the reward received at the time step $t+1$, and $s_t$ is the current state.

The latter are algorithms seeking to maximize the discounted cumulative reward, but directly learning and optimizing the parametrized policy $\pi(s|a;\theta)$~\citep{Sutton2018}.

A3C~\citep{Mnih2016} is a policy-based method, with the particularity that the \textit{actor} optimizes the policy, and, at the same time, the \textit{critic} estimates the state-value function. A3C allows for the deployment of several agents that run in parallel, each with a unique environment instance, and asynchronously, they update a shared global network. This guarantees stable and efficient learning, reducing the correlation between the experience gathered by the instantiated agents. As shown in~\citep{Gu2018}, ~\citep{Grondman2012}, and in ~\citep{Babaeizadeh2017}, A3C is a good alternative for discrete action spaces and high-dimensional state spaces.

Regarding the learning process, A3C uses the \textit{Advantage Function}, $A(s,a) = Q(s,a)-V(s)$, to measure the goodness of an action. This function can be approximated by estimating the value of the current state, the reward, and the discounted value of the next state. This is the \textit{Temporal Difference error}, or TD error at t:
\begin{equation}
\delta^{(t)} = r^{(t)} + \gamma V(s_{t+1}; \theta) - V(s_t; \theta)
\end{equation}
where $V(s_t; \theta)$ is the estimated value of $s_t$ given the policy parameters $\theta$. To smooth this estimation, we use the \textit{General Advantage Estimator (GAE)}:
\begin{equation}
A_{{GAE}}^{(t)} = \gamma \lambda A_{{GAE}}^{(t+1)} + \delta^{(t)}
\end{equation}
where $A_{{GAE}}^{(t)}$ is the GAE at time step $t$, and $\lambda$ is the trace decay factor that controls the bias-variance trade-off. On the other hand, the loss is defined as the sum of the \textit{Policy Loss} and the \textit{Value Loss}, defined as:
\begin{equation}
\text{P}_{\text{loss}} = -\sum_{i} \left( \log \pi(a_{ti} | s_{ti}; \theta) \cdot A_{\text{GAE}}^{(t)} - \beta H(\pi(s_t; \theta)) \right)
\end{equation}
where $i$ iterates over actions at $t$, $\pi(a_{ti} | s_{ti}; \theta)$ is the probability of taking $a_{ti}$, $\beta$ is the entropy regularization weight, and $H(\pi(s_t; \theta))$ is the policy entropy.
\begin{equation}
\text{V}_{\text{loss}} = \frac{1}{2} \sum_t \left( R_t - V(s_t; \theta) \right)^2
\end{equation}
where \(R_t\) is the n-step return at $t$.

\subsubsection{Proposed architecture}
\begin{figure}[h!]
\centering
\includegraphics[width=0.7\columnwidth]{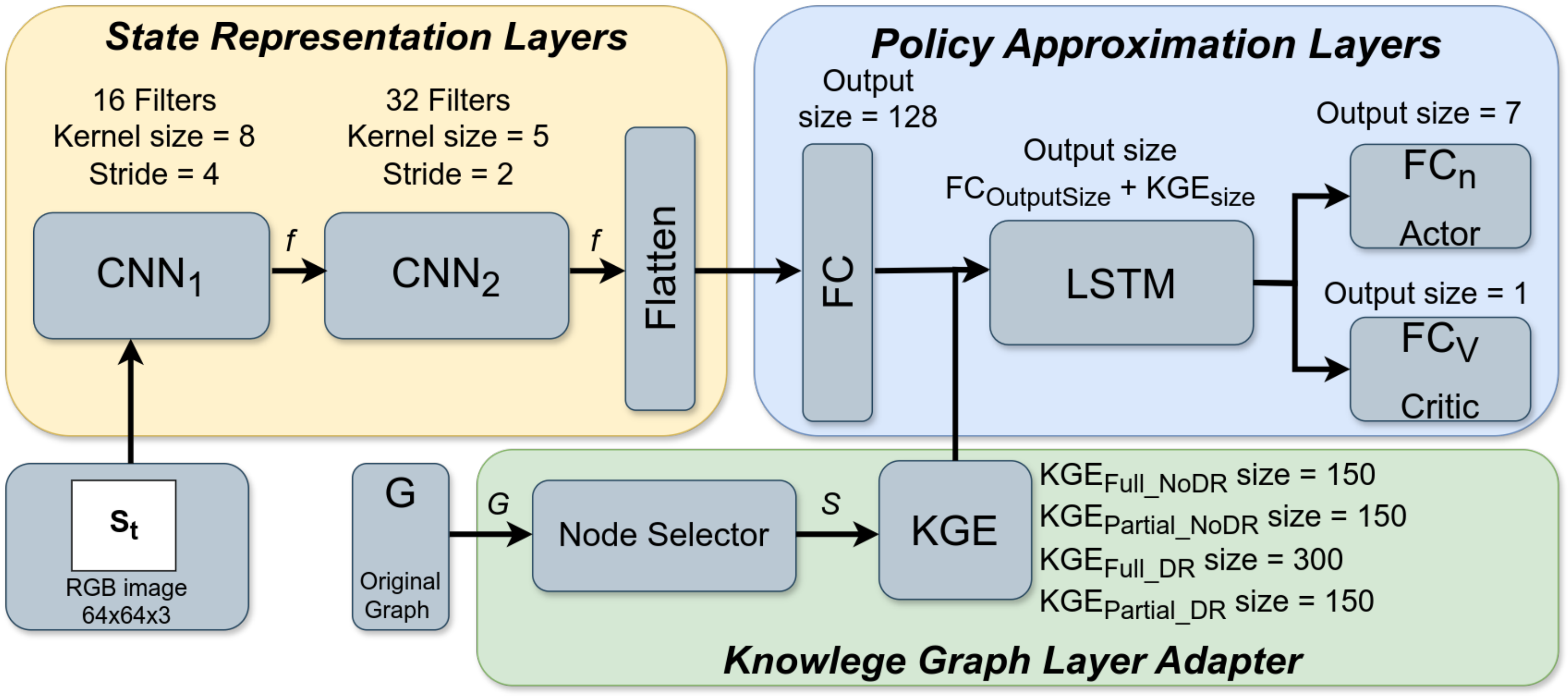}
\caption{Proposed A3C architecture to integrate KGEs in the DRL agent's learning process. The KGE size depends on the number of features encoded. The output is divided into seven actors for the TIAGo and six for the IRB120, plus the critic value.}
\label{fig_arch}
\end{figure}

Figure~\ref{fig_arch} shows the proposed architecture. The state representation layers are defined by a convolutional layer $\text{CNN}_1$ with 3$\times$3 kernel and stride 4 and another convolutional layer $\text{CNN}_2$ with 5$\times$5 kernel and stride 2. Both layers have at the end a non-linear $f$ implemented through the ReLU function, $f(x)=max(0,x)$. The policy approximation block is formed by a 1152$\times$128 fully connected layer (FC), a Long-Short Term Memory (LSTM), whose capacity depends on the size of the integrated KGE. Since the agent commands the orientation of each robot joint, the A3C actor layers are $n$ FC layers, where $n$ is the number of robot joints, i.e., 7 for the TIAGo and 6 for the IRB120. The dimension of the $n$ FC blocks is determined by the discrete set of actions, which has seven actions per robot joint (Table~\ref{tab:MDP_def}). We use the softmax function to convert the activations into probabilities. Therefore, the actor outputs are $n$ 7-dimension arrays that contain the probability associated with each action. Finally, the A3C critic is a single FC layer that estimates the value of $V(s)$. We will refer to the Baseline Agent or Baseline Model (BM) as the agent trained under this architecture with just the RGB image. It will serve as a reference to quantify the improvement obtained by integrating the KGE.

The originality of our proposal lies in the integration of embeddings coming from a graph containing some environment semantic information. This new input, which does not change throughout the training, might be placed next to the policy approximation blocks, since it should complement the state representation from the ``vision'' layers. Therefore, the most appropriate place to integrate the KGE is before the LSTM layer, linking the embeddings to the activations coming from the FC. The FC layer receives the representation coming from the CNNs and provides a higher abstraction level. The size of the embeddings and, consequently, the size of the LSTM will depend on the case study. In general, the embedding concatenated is a 150-dimension array, except for the one used when DR is applied to the targets' color, which is a 300-dimension array. As a result, the LSTM size will be 128, when no KGE is integrated, 428 when the KGE includes all the information from the randomized targets, which is an assumable size for our architecture, and 278 otherwise.

\subsubsection{Training procedure}
We train the models for 70~million steps using the A3C algorithm. We use a random orthogonal initialization for all the layers. Based on the computational resources available, the number of asynchronous agents instantiated is 17 plus one for the interim evaluation. This interim evaluation happens every 50,000~steps during training, and it assesses the global shared model for 40~episodes. These episodes keep the same 40 initial configurations, i.e., robot and target poses, across all the interim evaluations. At the beginning of each episode, a target is selected and placed randomly within the workspace area. The non-selected targets do not appear on the image. The rewarding distance is \SI{5}{cm}, and the allowed orientation deviation is $\pm15^o$. 

The optimizer selected was the Root Mean Square Propagation (RMSprop)~\citep{Tieleman2012}. The values of the most relevant hyperparameters are learning rate $1e^{-4}$, discount factor ($\gamma$) 0.99, entropy weight ($\beta$) 0.01, trace decay ($\lambda$) 1, and RMSprop decay 0.99.

\subsubsection{Post-training evaluation procedure}
After the training process, we select the best model according to the average return obtained in the interim evaluations. Then, we perform a post-training evaluation across 1,000 episodes with a different seed and initial robot and target poses from the ones used during training. The rewarding distance and allowed orientation deviation in the post-training evaluation are \SI{10}{cm} and $\pm17^o$, respectively. We modify these training constraints because we believe that training in more restrictive circumstances yields better post-training results. In the assessment, we calculate the mean and standard deviation (Std) return, the mean and Std episode length, the mean, Std, and maximum failure distance, and the accuracy. In our context, we understand accuracy as the percentage of successful episodes where the robot reached the target within the required distance. We formulate the accuracy as:
\begin{equation}
\label{eq:Accuracy}
\text{Accuracy (\%)} = \frac{\sum_{i=1}^{N} \mathbb{I}(\text{dist}_{i} \leq \SI{10}{cm})}{N} \times 100
\end{equation}
where $N$ is the total number of evaluated episodes, $dist_{i}$ is the relative distance between the gripper and the target in the $i^{th}$ episode, and $\mathbb{I}$ is the indicator function.

\subsubsection{Experiment description}
To analyze the influence of KGE in the learning process of a DRL agent, we present two sets of experiments. We first train three agents in an environment where each target has one possible color. The difference between these agents is the information we provide through the KGE. The BM agent has no KGE integrated. Then, there is an agent with only partial knowledge about the targets, just the object type (partial KGE). Finally, there is an agent with a complete description of the target type and color (full KGE). 

In the second set of experiments, we decided to increase the complexity of the problem by introducing DR to the target color. Each object has two possible colors (Section~\ref{sec:EnvironmentSetup}) selected randomly at the episode beginning after the target choice and its placement. As in the first set, we train three different agents. A BM agent with no KGE, an agent with the partial KGE that contains only the object type, and an agent with the full KGE that has the object type and the possible colors per object.

We complement both groups of experiments with a quantitative and qualitative analysis of the joints' angles distribution to research how the exploration and exploitation capabilities of the agents might be affected by the KGEs.

%%%%%%%%%%%%%%%%%%%%%%%%%%%%%%%%%%%%%%%%%%
\section{Results}
\label{sec:Results}
\subsection{Experiments without DR}
\begin{figure}[h!]
\centering
\includegraphics[width=0.5\columnwidth]{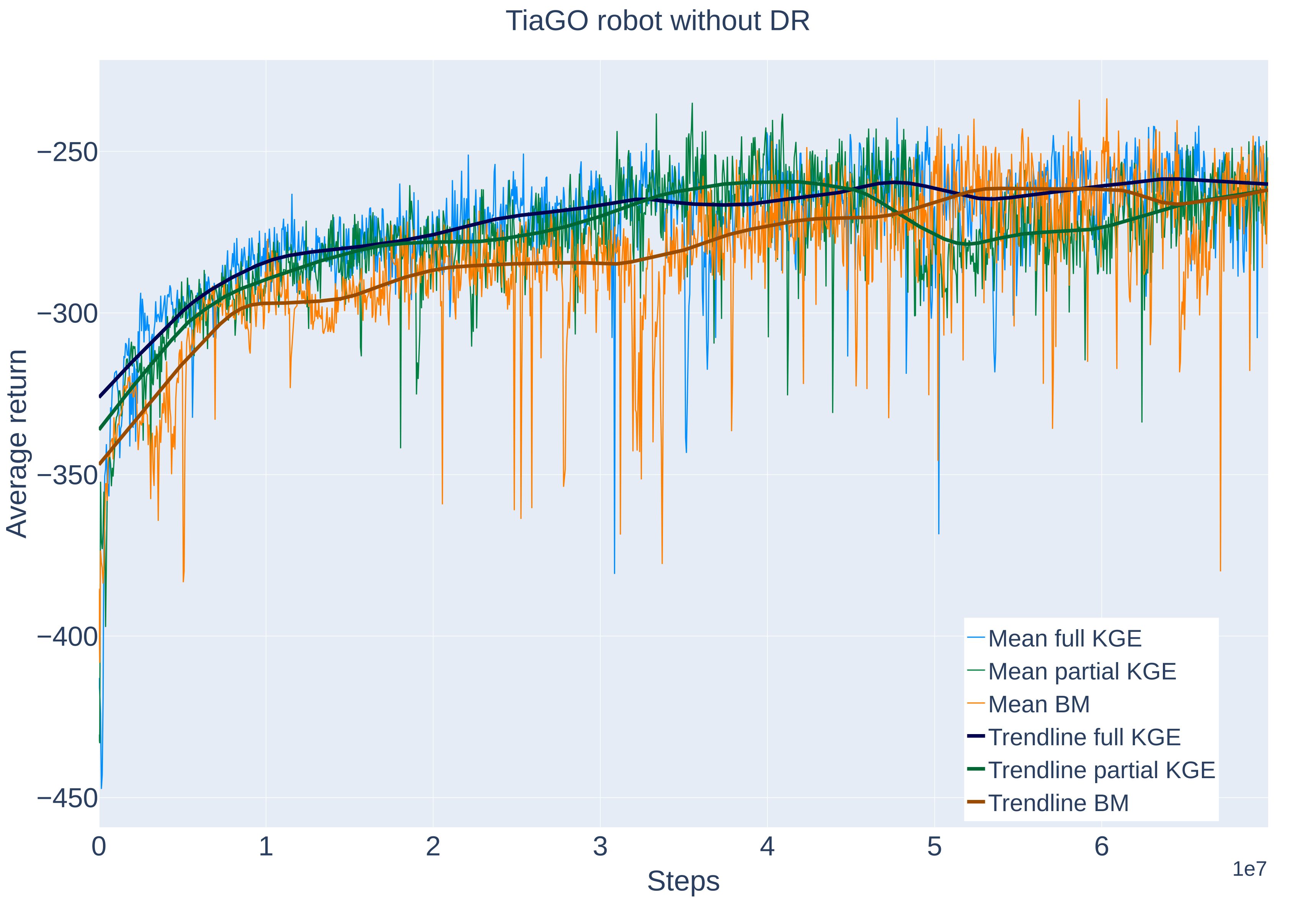}
\caption{Training curves of the agents trained with the TIAGo and fixed targets' colors. The blue curve is the full KGE model, the green is the partial KGE, and the orange is the BM. The training process lasts for 70~M steps.}
\label{fig_training_NoDR_tiago}
\end{figure}

\begin{figure}[h!]
\centering
\includegraphics[width=0.5\columnwidth]{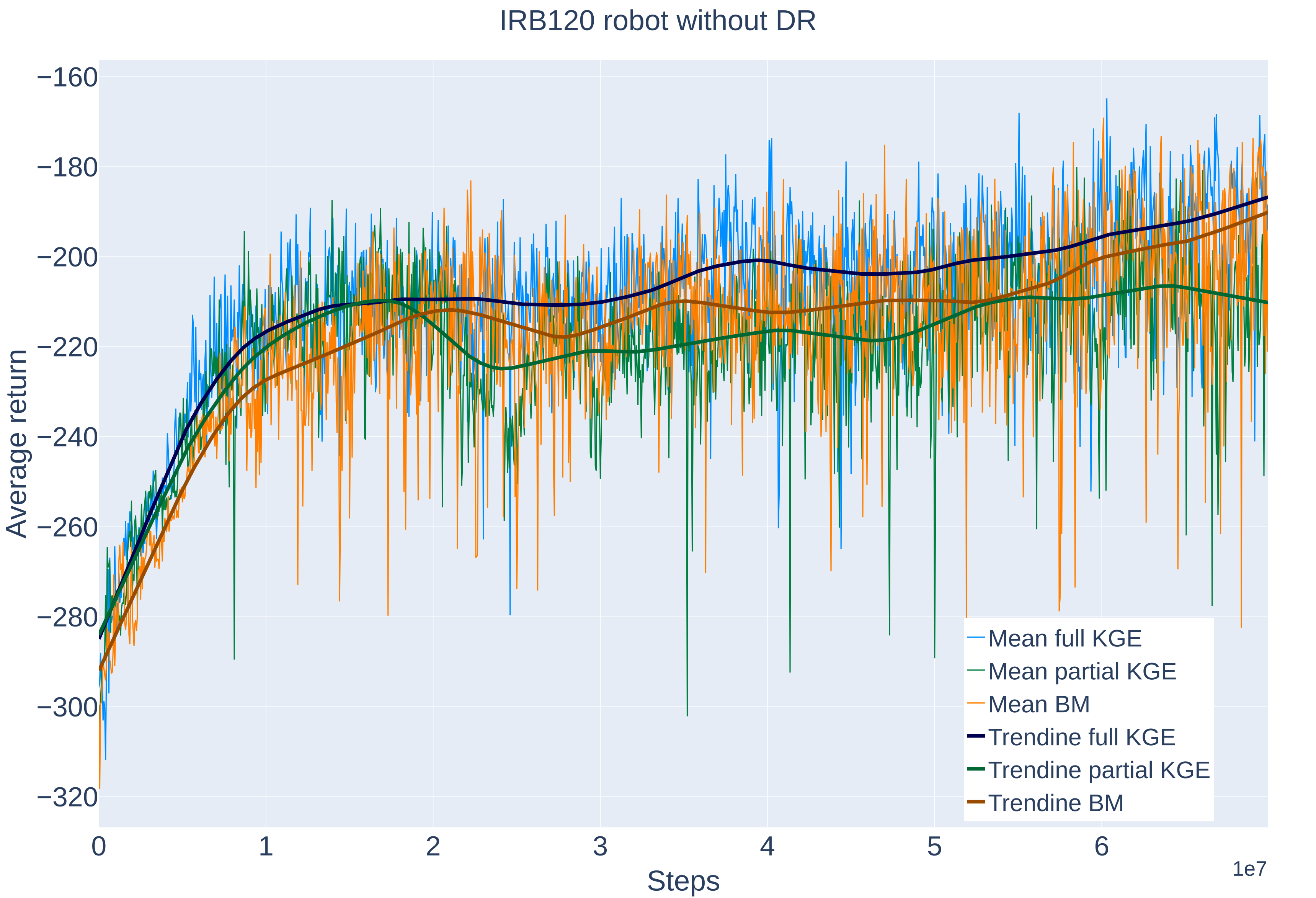}
\caption{Training curves of the agents trained with the IRB120 and fixed targets' colors. The blue curve is the full KGE model, the green is the partial KGE, and the orange is the BM. The training process lasts for 70~M steps.}
\label{fig_training_NoDR_irb120}
\end{figure}

Figure~\ref{fig_training_NoDR_tiago} and Figure~\ref{fig_training_NoDR_irb120} show the training results in the environment with invariant features for the TIAGo and IRB120 robots, respectively. In the case of TIAGo, Figure~\ref{fig_training_NoDR_tiago} shows that the agent trained with the complete KGE learns faster and remains with a higher return than the agent trained with the partial KGE. In the first case, the best model is achieved at 48~M steps and has \qty{72}{\percent} accuracy, whereas in the second, it is at 36~M steps and has \qty{70}{\percent} accuracy. On the other hand, the differences with the BM in terms of learning speed and accuracy are higher. Although both agents reach the same steady regime, the best BM is at 60~M steps, and its accuracy is \qty{60}{\percent}. Hence, the improvement gain using KGE under a setup with invariant features is 12 percentual points.

Regarding the IRB120 robot, Figure~\ref{fig_training_NoDR_irb120} shows that, although there is a larger difference between the full KGE and the partial KGE trendlines, the accuracy of the best models are \qty{92}{\percent} and \qty{91}{\percent}, respectively. They are obtained in the 60~M step and the 59~M step. On the other hand, comparing the results with the BM, we notice that its learning is slower than the agent with the full KGE, which stands above the BM during the process. The performance of the BM best agent, which occurs in the 60~M step, reaches \qty{80}{\percent} accuracy. Therefore, the accuracy improvement in this setup without DR is also around \qty{12}{\percent} when using the KGE as an additional input to the model.

\begin{figure}[h!]
\centering
\includegraphics[width=0.8\columnwidth]{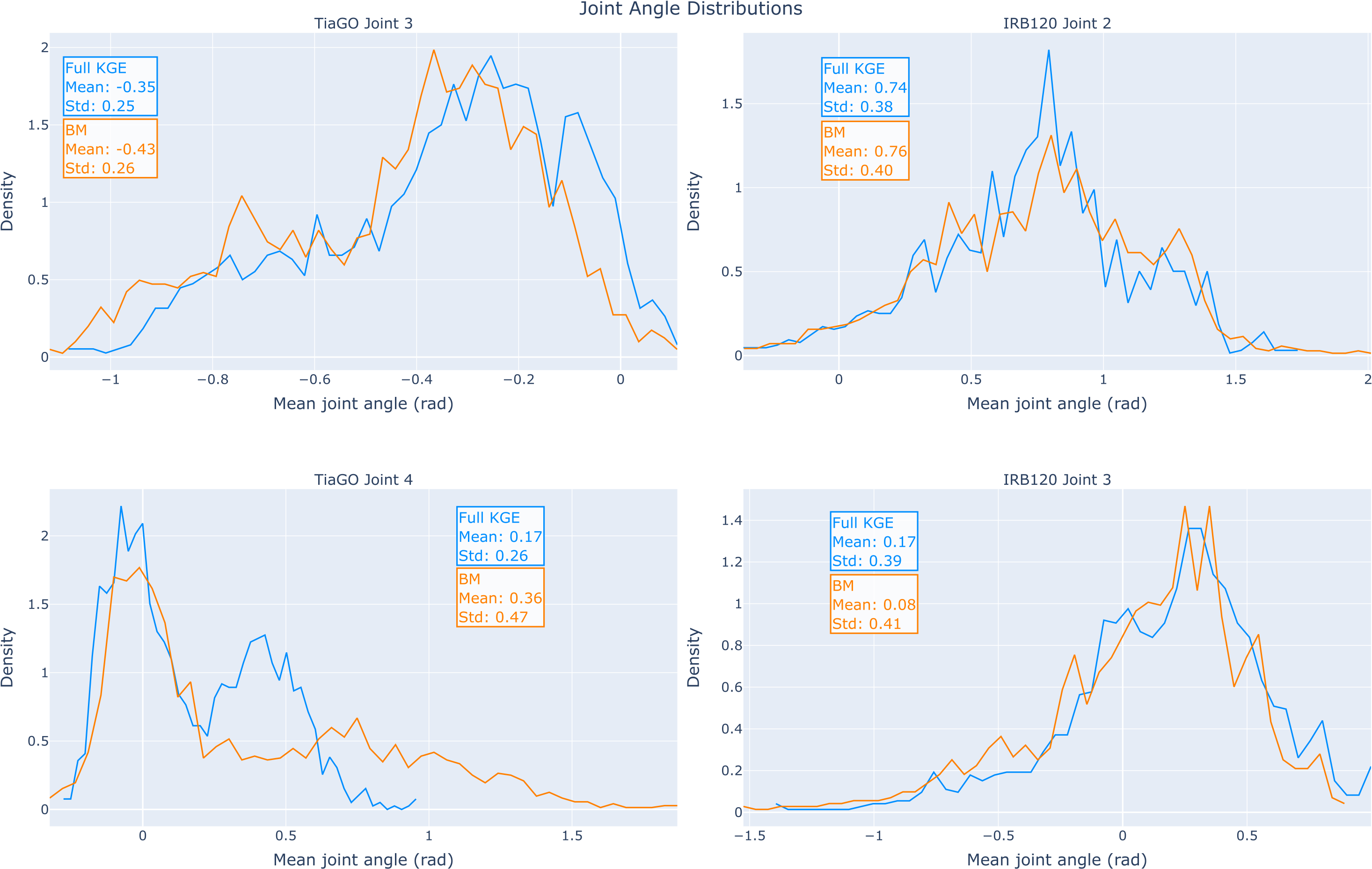}
\caption{Joint angle distributions for the best full KGE agent (blue), and the best BM agent (orange), without DR. The first column corresponds to the TIAGo and the second to the IRB120. The selected joints for TIAGo are joints 3 and 4, and for the IRB120, they are joints 2 and 3.}
\label{fig_jointAngleDistrib_noDR}
\end{figure}

Lastly, the plots in the first column of Figure~\ref{fig_jointAngleDistrib_noDR} show the joints' angles distribution for the TIAGo, while the Figure~\ref{fig_jointAngleDistrib_noDR} second column graphs represent the IRB120 joints. We compare the distributions of the full KGE agents, represented in blue, with the BMs, shown in orange. The selected joints for the TIAGo are joints 3 and 4, and for the IRB120, they are joints 2 and 3. These joints were selected because they are the ones that present meaningful differences in the distribution of their values, which is interpreted as an improvement in the agent's ability to learn more straightforward the right policy. TIAGo's joint 3 experienced a slight reduction in the standard deviation of 0.01 points, but a significant mean variation. TIAGo's joint 4, however, has a large standard deviation decrease of 0.21 points and a mean shift. Besides, the distribution changes to a bimodal shape, which might indicate some change in the agent's policy. In the case of IRB120, joint 2 has a standard deviation decrease of 0.02 points, as well as joint 3, which also presents a relevant mean difference in the distributions. 

\subsection{Experiments with DR}
\begin{figure}[h!]
\centering
\includegraphics[width=0.5\columnwidth]{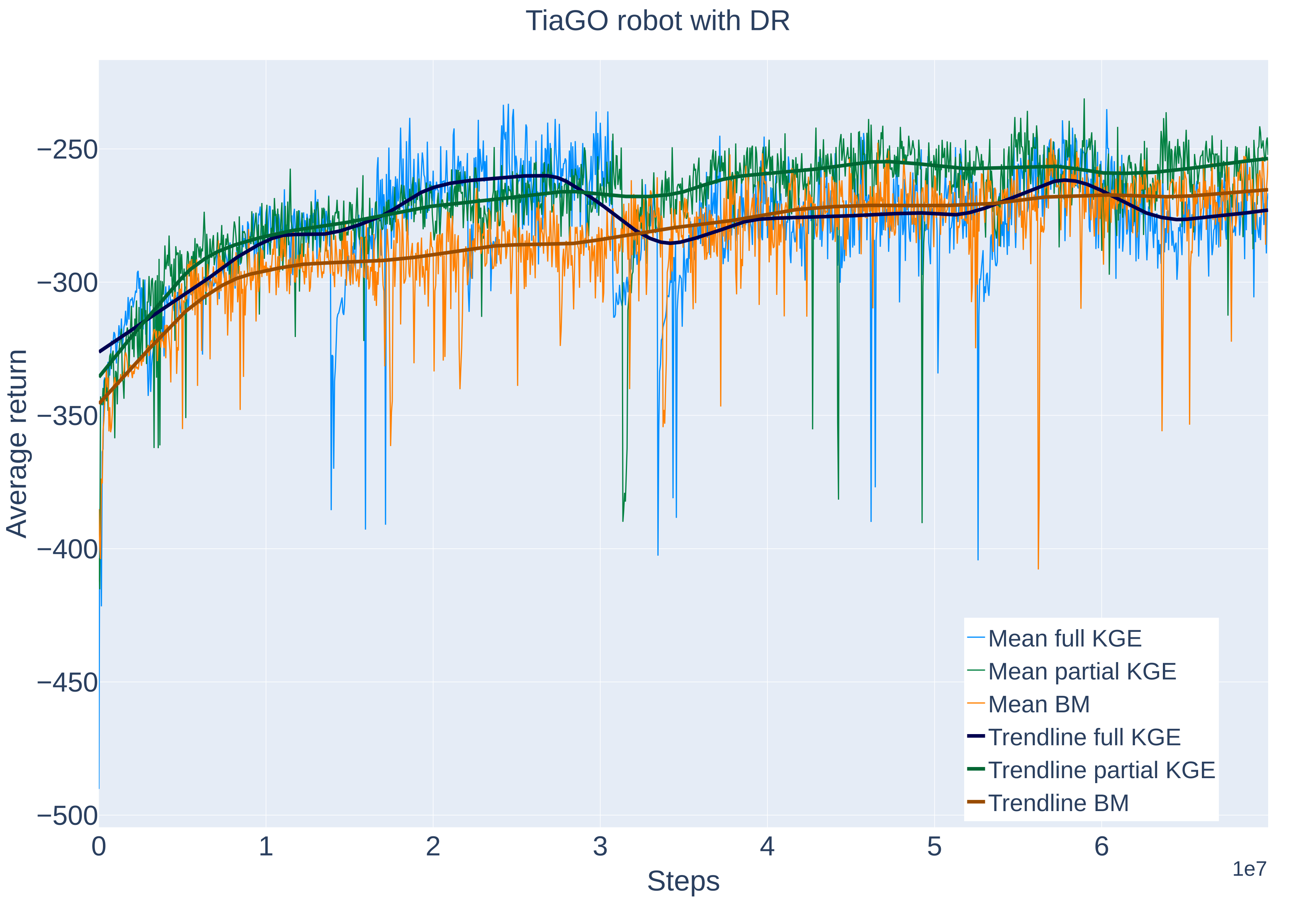}
\caption{Training curves of the agents trained with the TIAGo and DR on the targets' colors. The blue curve is the full KGE model, the green is the partial KGE, and the orange is the BM. The training process lasts for 70~M steps.}
\label{fig_training_DR_tiago}
\end{figure}

\begin{figure}[h!]
\centering
\includegraphics[width=0.5\columnwidth]{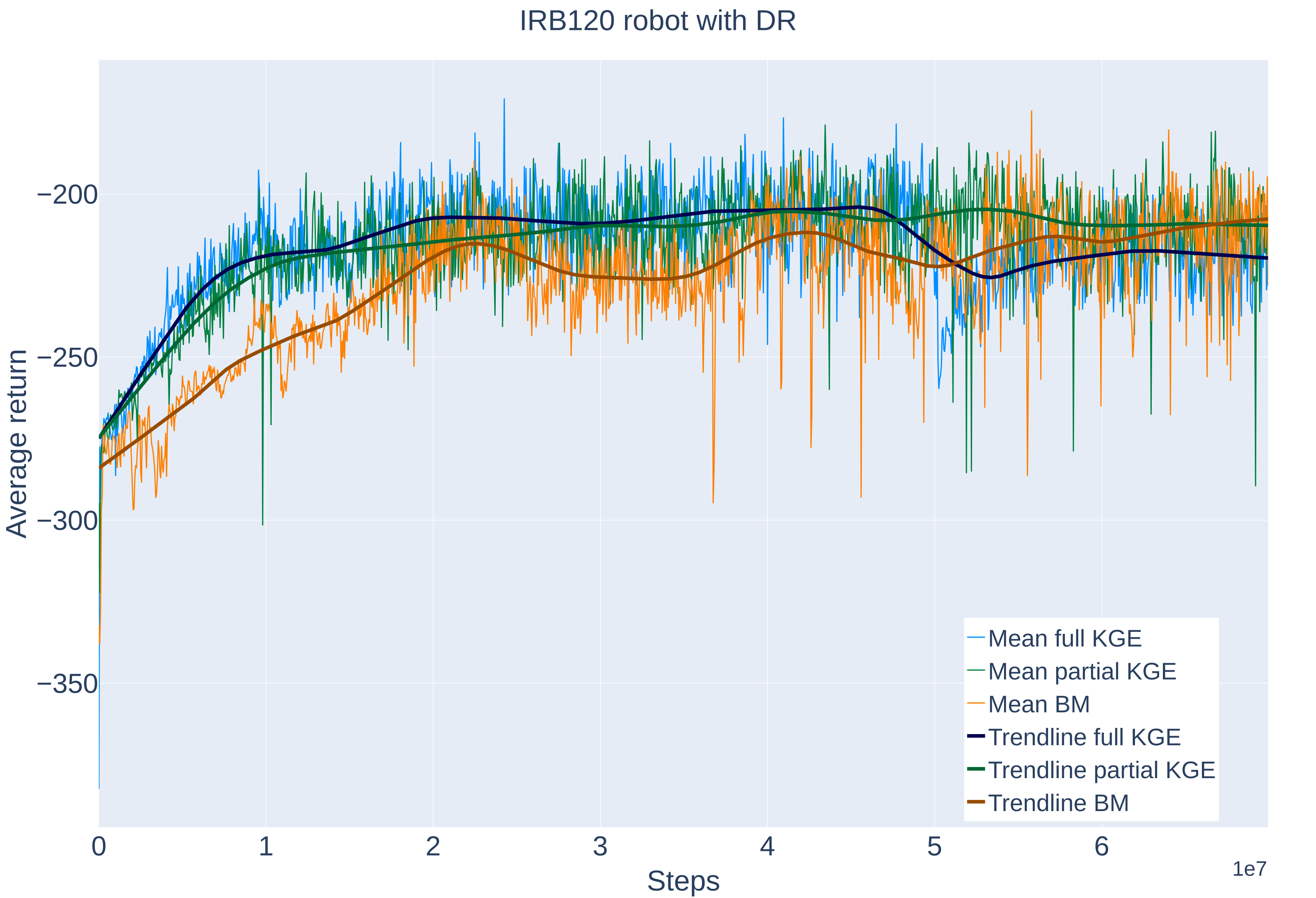}
\caption{Training curves of the agents trained with the IRB120 and DR on the targets' colors. The blue curve is the full KGE model, the green is the partial KGE, and the orange is the BM. The training process lasts for 70~M steps.}
\label{fig_training_DR_irb120}
\end{figure}

Figure~\ref{fig_training_DR_tiago} and Figure~\ref{fig_training_DR_irb120} present the training curves for the three models trained in the TIAGo and IRB120 setups when the targets' color change along the training episodes. For the first scenario, we spot that the best model of the agent with the full KGE achieves \qty{79}{\percent} accuracy, higher than the \qty{62}{\percent} obtained by the agent with the partial KGE, and the \qty{63}{\percent} of the BM. Additionally, we can see that the learning is faster, and as a result, the best model is obtained at the 24~M step, earlier than in the partial KGE model, which happens at the 59~M step and in the BM at the 57~M step. Therefore, we see that the overall improvement of using a complete KGE with the target type and color in a setup where this last feature can change is around \qty{16}{\percent} with respect to the BM, and it is achieved in half of the training time.

In the scenario with the IRB120 robot, the agent with the complete KGE learns faster and yields higher than the BM. Whereas the first one reaches \qty{86}{\percent} accuracy in the 24~M step, the latter attains \qty{76}{\percent} in the 56~M step. If we compare the results with the model with partial KGE, it obtains \qty{87}{\percent} accuracy in the 43~M step, a comparable success rate. Table~\ref{tab:experiments_results} collects the accuracy results of the trained agents. 

\begin{table}[h!]
\centering
\caption{Experiments made to analyze the KGE influence in the DRL agent learning process, using TIAGo and IRB120 robotic arms. There are two sets: without DR and with DR. Each set includes three types of agents: BM, Partial KGE, and Full KGE.}
\resizebox{0.7\columnwidth}{!}{
\begin{tabular}{c c c c c}
    \textbf{Experiment Set} & \textbf{Robots} & \textbf{Agent Type} & \textbf{Accuracy \%} & \textbf{Best model step} \\
    \midrule\midrule
    \multirow{6}{*}{Without DR} & \multirow{3}{*}{TIAGo} & BM & \qty{60}{\percent} & 60~M\\
                                & & Partial KGE & \qty{70}{\percent} & 36~M\\
                                & & Full KGE & \qty{72}{\percent} & 48~M\\
    \cmidrule{2-5}
                                & \multirow{3}{*}{IRB120} & BM & \qty{80}{\percent} & 60~M \\
                                & & Partial KGE & \qty{91}{\percent} & 59~M\\
                                & & Full KGE & \qty{92}{\percent} & 60~M\\
    \midrule
    \multirow{6}{*}{With DR} & \multirow{3}{*}{TIAGo} & BM & \qty{63}{\percent} & 57~M\\
                             & & Partial KGE & \qty{62}{\percent} & 59~M \\
                             & & Full KGE & \qty{79}{\percent} & 24~M \\
    \cmidrule{2-5}
                             & \multirow{3}{*}{IRB120} & BM & \qty{76}{\percent} & 56~M\\
                             & & Partial KGE & \qty{87}{\percent} & 43~M\\
                             & & Full KGE & \qty{86}{\percent} & 24~M\\
    \bottomrule
\end{tabular}}
\label{tab:experiments_results}
\end{table}

\begin{figure}[h!]
\centering
\includegraphics[width=0.8\columnwidth]{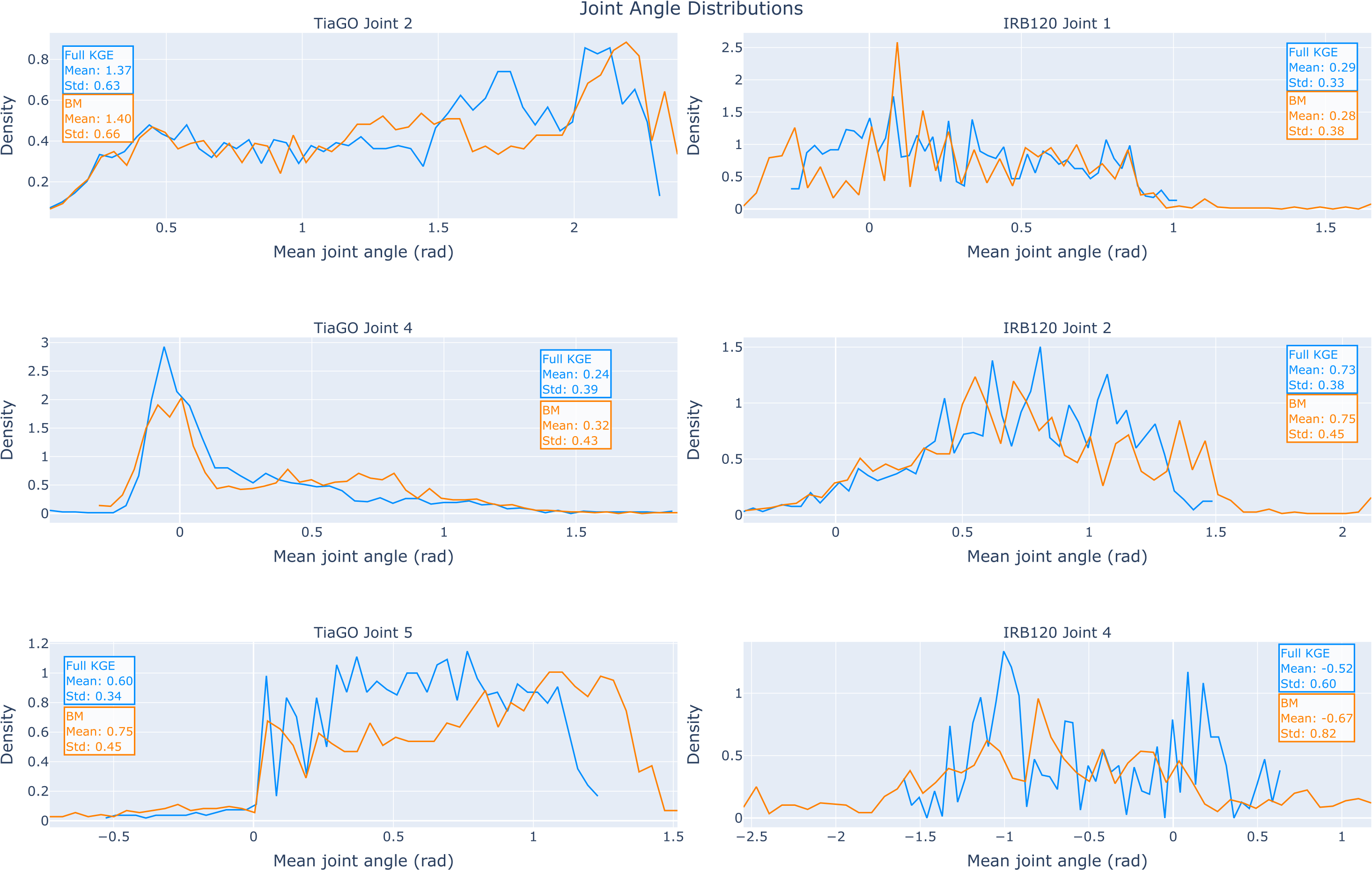}
\caption{Joint angle distributions for the best full KGE agent (blue) and the best BM agent (orange), with DR. The first column corresponds to the TIAGo and the second to the IRB120. The selected joints for TIAGo are joints 2, 4, and 5, and for the IRB120, they are joints 1, 2, and 4.}
\label{fig_jointAngleDistrib_DR}
\end{figure}

Figure~\ref{fig_jointAngleDistrib_DR} exhibits the joints' angle distributions for the TIAGo and IRB120 models with DR. For the TIAGo, the most significant results appear on joints 2, 4, and 5. They are displayed in Figure~\ref{fig_jointAngleDistrib_DR} first column. In all of them, the full KGE agent, the blue curve, presents lower standard deviation values on the joints' distributions than the BM, the orange curve. In the case of joint 2, it decreases 0.03~points, 0.04~points in joint 4, and 0.11~points in joint 5. For the IRB120, we analyze joints 1, 2, and 4, shown in Figure~\ref{fig_jointAngleDistrib_DR} second column. As in the TIAGo, the agent with the complete KGE, the blue plot, has less standard deviation than the BM, the orange curve. This reduction is 0.05~points in joint 1, 0.07~points in joint 2, and 0.12~points in joint 4. We decided not to include the result of the other joints because it does not have a significant change. 

\begin{figure}[H]
    \centering
    \begin{minipage}{0.25\linewidth}
        \centering
        \includegraphics[width=\linewidth]{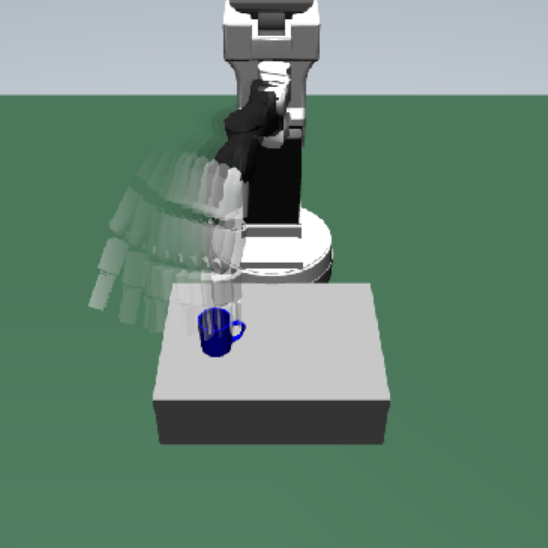}
        \vspace{1mm}
        \text{(a)}
    \end{minipage}
    \hspace{0.05\linewidth}
    \begin{minipage}{0.25\linewidth}
        \centering
        \includegraphics[width=\linewidth]{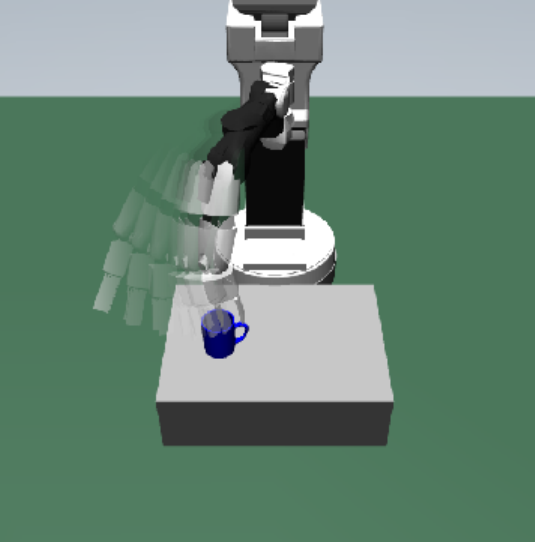}
        \vspace{1mm}
        \text{(b)}
    \end{minipage}
    \caption{TIAGo initial movements in the BM agent (a) and the full KGE agent (b) when DR is applied in the environment.}
    \label{fig_TiaGOmov_DR}
\end{figure}

\begin{figure}[H]
    \centering
    % Primera fila
    \begin{minipage}{0.11\linewidth}
        \centering
        \includegraphics[width=\linewidth]{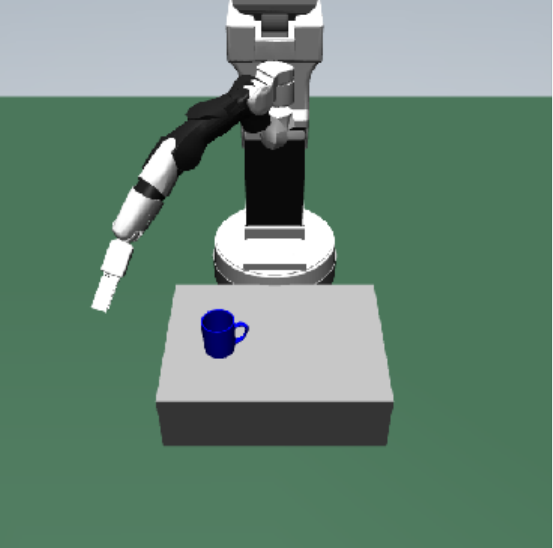}
        \text{step 1}
    \end{minipage}
    \begin{minipage}{0.11\linewidth}
        \centering
        \includegraphics[width=\linewidth]{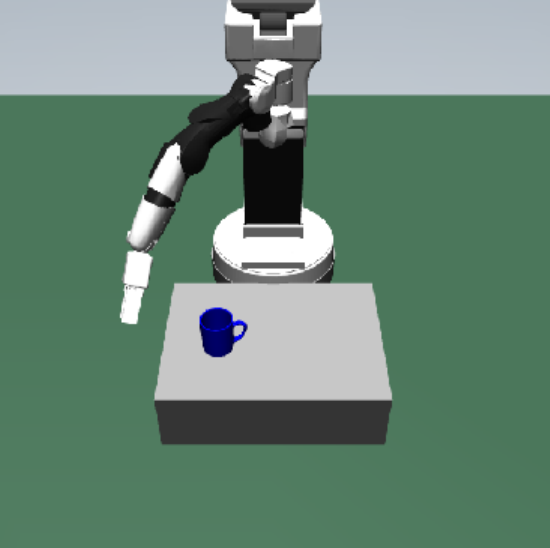}
        \text{step 2}
    \end{minipage}
    \begin{minipage}{0.11\linewidth}
        \centering
        \includegraphics[width=\linewidth]{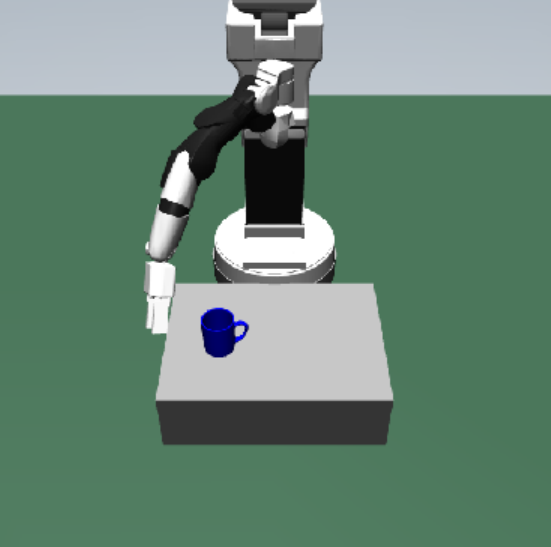}
        \text{step 3}
    \end{minipage}
    \begin{minipage}{0.11\linewidth}
        \centering
        \includegraphics[width=\linewidth]{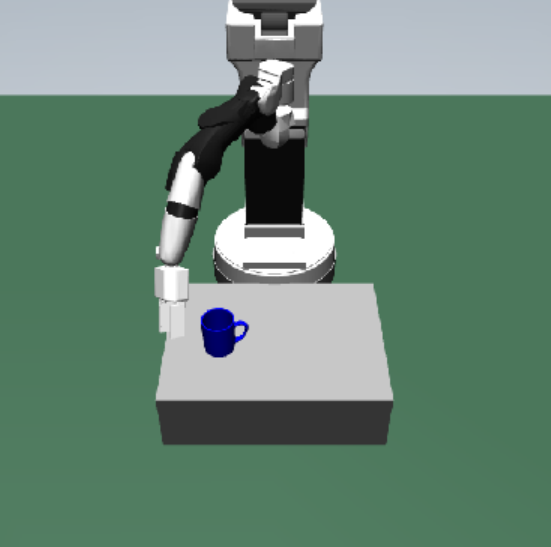}
        \text{step 4}
    \end{minipage}
    \begin{minipage}{0.11\linewidth}
        \centering
        \includegraphics[width=\linewidth]{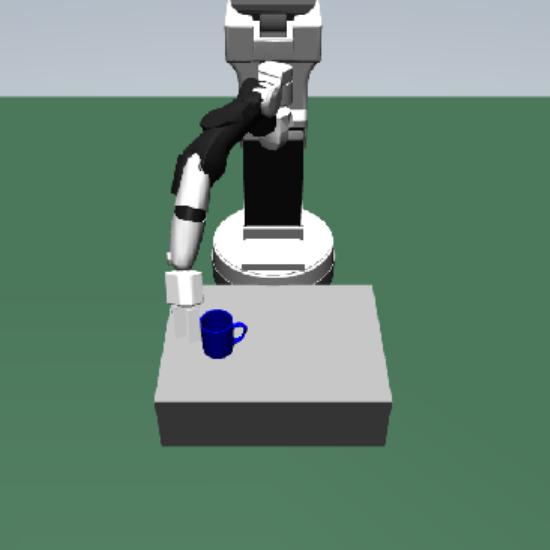}
        \text{step 5}
    \end{minipage}
    \begin{minipage}{0.11\linewidth}
        \centering
        \includegraphics[width=\linewidth]{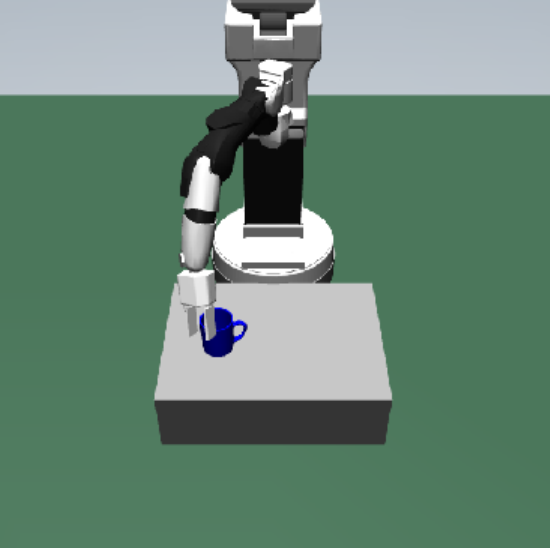}
        \text{step 6}
    \end{minipage}
    \begin{minipage}{0.11\linewidth}
        \centering
        \includegraphics[width=\linewidth]{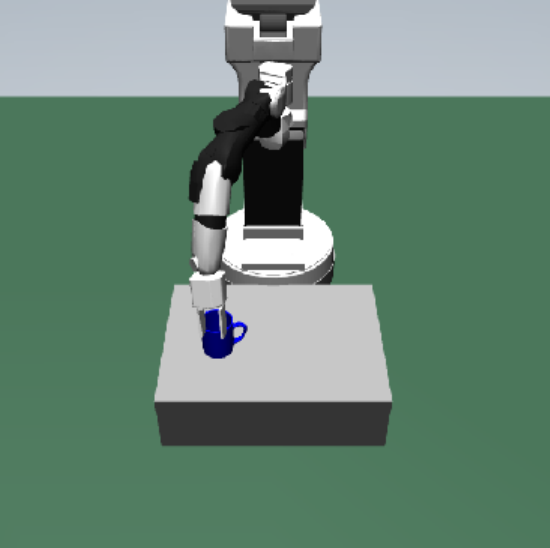}
        \text{step 7}
    \end{minipage}
    \begin{minipage}{0.11\linewidth}
        \centering
        \includegraphics[width=\linewidth]{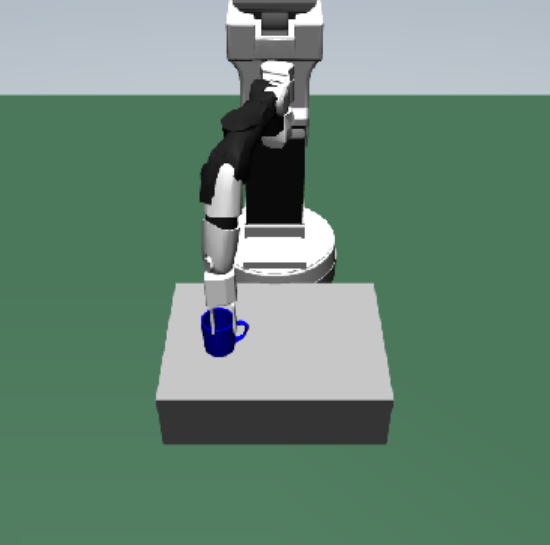}
        \text{step 8}
    \end{minipage}

    % Segunda fila
    \begin{minipage}{0.11\linewidth}
        \centering
        \includegraphics[width=\linewidth]{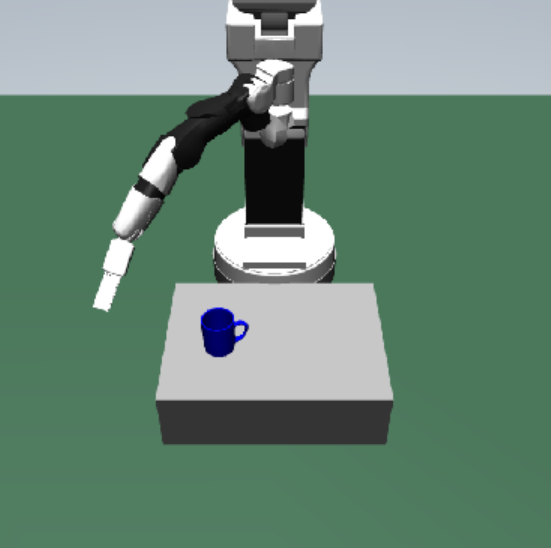}
    \end{minipage}
    \begin{minipage}{0.11\linewidth}
        \centering
        \includegraphics[width=\linewidth]{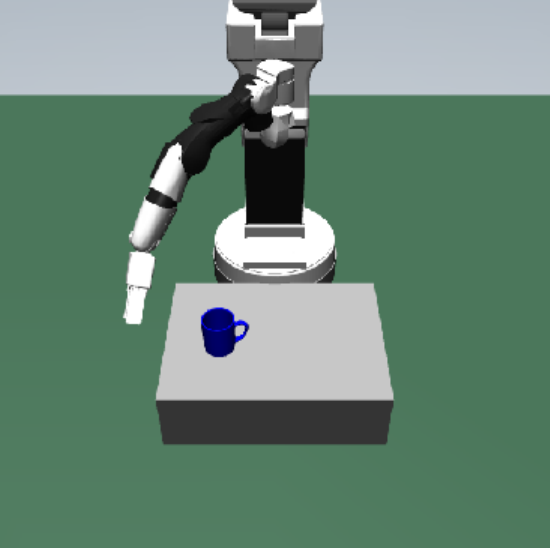}
    \end{minipage}
    \begin{minipage}{0.11\linewidth}
        \centering
        \includegraphics[width=\linewidth]{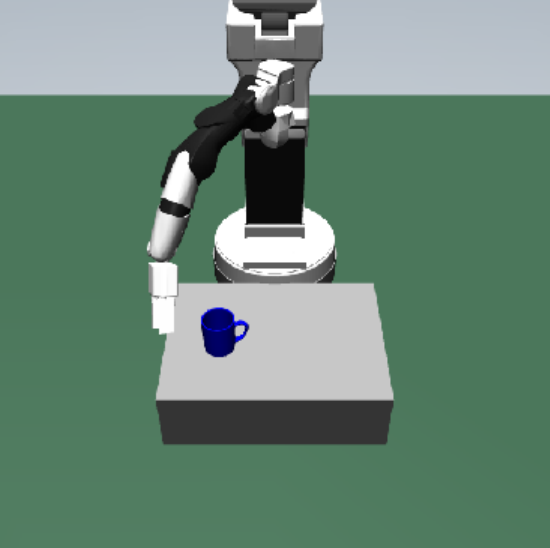}
    \end{minipage}
    \begin{minipage}{0.11\linewidth}
        \centering
        \includegraphics[width=\linewidth]{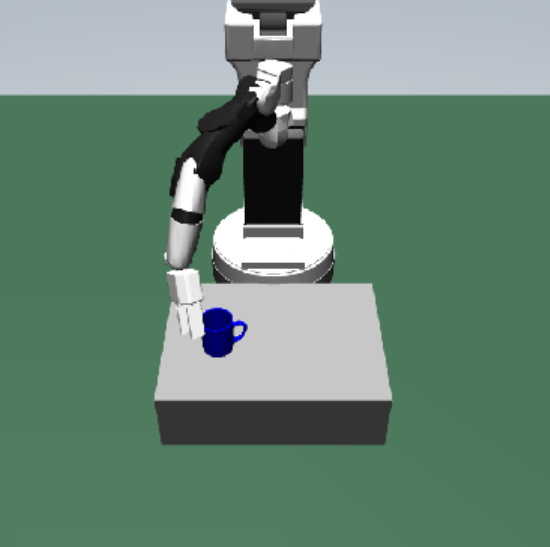}
    \end{minipage}
    \begin{minipage}{0.11\linewidth}
        \centering
        \includegraphics[width=\linewidth]{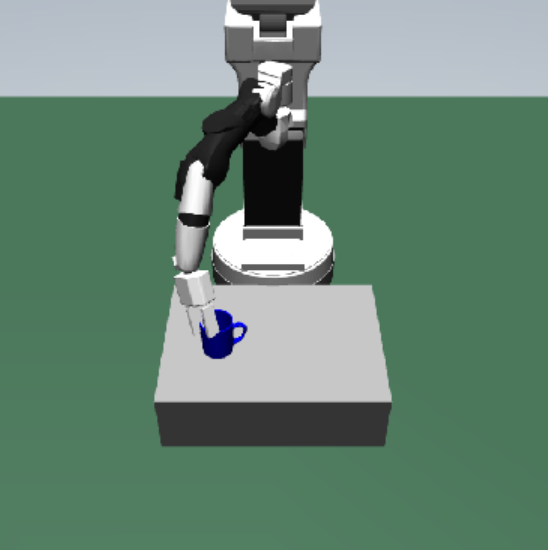}
    \end{minipage}
    \begin{minipage}{0.11\linewidth}
        \centering
        \includegraphics[width=\linewidth]{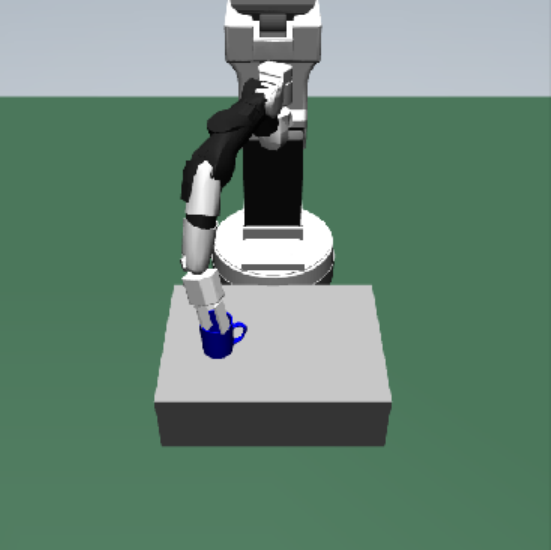}
    \end{minipage}
    \begin{minipage}{0.11\linewidth}
        \centering
        \includegraphics[width=\linewidth]{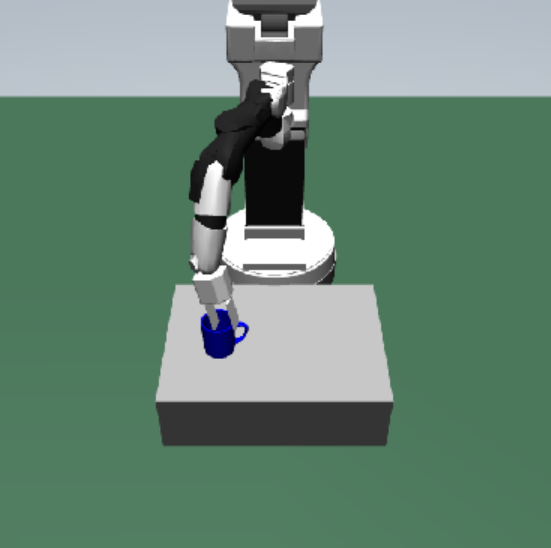}
    \end{minipage}
    \begin{minipage}{0.11\linewidth}
        \centering
        \includegraphics[width=\linewidth]{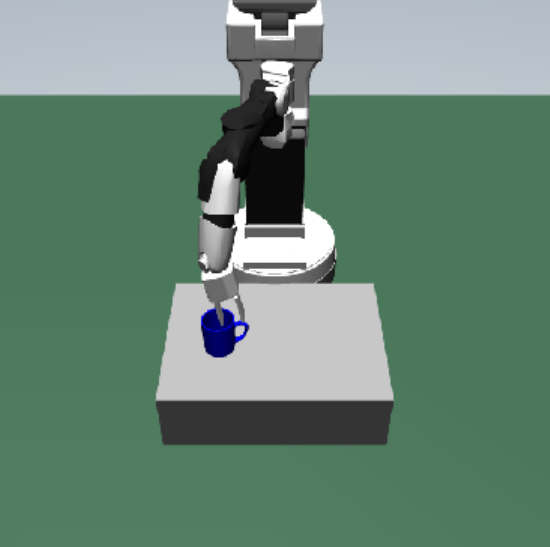}
    \end{minipage}
    \caption{TIAGo initial movements in the BM agent (first row) and the full KGE agent (second row) when DR is applied in the environment. During the first steps of the episode, the KGE agent needs fewer steps to approach the object and reach a more suitable affordance orientation. Both episodes finished successfully, the BM agent in 34 steps and the full KGE agent in 20 steps.}
    \label{fig_TiaGOmov_DR_actions}
\end{figure}

Figure~\ref{fig_TiaGOmov_DR} and Figure~\ref{fig_TiaGOmov_DR_actions} illustrate the differences commented about the joint angle distribution depending on the presence or absence of semantic knowledge. As shown, the BM agent requires more intermediate movements from the beginning of the episode to reach the intermediate position where fine orientation and approach adjustments are made to grasp the target. In contrast, the agent equipped with the complete KGE typically reaches the grasping point more directly and with a more favorable pre-alignment, allowing for a smoother and more efficient approach. As Figure~\ref{fig_TiaGOmov_DR_actions} shows, in steps 4 and 5 of the episode, the KGE agent has already reached the mug, and it is rotating the gripper to align it with the affordance point. Both episodes ended up successfully, but the BM agent needed 34 steps, while the full KGE agent needed only 20 steps.

\subsection{Robustness of the results}
To statistically assess whether the observed differences in performance across models were significant, we conducted a one-way Analysis of Variance (ANOVA) on the success rates of each pair of models: TIAGo/IRB120 with DR or without DR for the full KGE agent and the BM agent. Each model was evaluated over 30 independent runs, each initialized with a different random seed, and each run consisted of 100 episodes. The null hypothesis ($H_0$) assumes that the mean success rates of the two models being compared are equal, i.e., any observed differences are due to random variation.

The results are gathered in Table~\ref{tab:anova_results}. For the TIAGo robot, in the environment where DR is applied, the success rate of the full KGE agent was significantly higher than for the BM agent. Similarly, in the standard setting, the full KGE agent outperformed BM with statistical significance. For the IRB120 robot, the differences were also pronounced: under DR, the full KGE agent achieved significantly higher success rates than BM agent, and in the standard setting, the KGE agent again significantly outperformed the BM agent. These results confirm that the proposed KGE models consistently achieve superior performance compared to the BM across both robots and training conditions, with statistically significant margins. The high $F$-values and extremely low $p$-values indicate that the between-group variance greatly exceeds the within-group variance, supporting the robustness of the observed performance differences and justifying the rejection of the null hypothesis in all cases.

\begin{table}[h!]
\centering
\caption{One-way ANOVA results comparing KGE and BM agents for each robot and training setting (with and without domain randomization). Each result is based on 30 evaluations with different seeds, each consisting of 100 episodes.}
\resizebox{0.55\columnwidth}{!}{
\begin{tabular}{c c c c}
    \textbf{Experiment Set} & \textbf{Robot} & \textbf{F-value} & \textbf{p-value} \\
    \midrule\midrule
    \multirow{2}{*}{Without DR} 
        & TIAGo   & 16.47   & $< 0.001$ \\
        & IRB120  & 150.88  & $< 1 \times 10^{-17}$ \\
    \midrule
    \multirow{2}{*}{With DR} 
        & TIAGo   & 247.14  & $< 1 \times 10^{-21}$ \\
        & IRB120  & 111.08  & $< 1 \times 10^{-14}$ \\
    \bottomrule
\end{tabular}}
\label{tab:anova_results}
\end{table}

%%%%%%%%%%%%%%%%%%%%%%%%%%%%%%%%%%%%%%%%%%
\section{Discussion}
\label{sec:Discussion}
According to the results presented, we can say that, in general, the TIAGo's agents achieve less accuracy than IRB120's because their environment is more challenging. We argue that the morphology of the arm, the fact that it has one more DoF than the IRB120, its reach, and its color are factors that increase the complexity of the problem, although the task is the same. 

In the experiments carried out with the TIAGo and with the IRB120 without DR we consistently observed an improvement in the transitory regime and a decrease in the learning process time. For the TIAGo, there is an improvement of \qty{25}{\percent} in the learning time between the full KGE agent and the BM. In the IRB120, although the best models are achieved at the same time, the full KGE trendline is always above the BM, and its transitory is much faster. With respect to the success rates, in both setups, there is almost no difference between the full and the partial KGE. This might prove that, since the color is an invariant feature in these experiments, the presence of this specification on the embeddings does not change the agent performance significantly. However, we do obtain \qty{12}{\percent} accuracy improvement with respect to the BM when we evaluate the full KGE agent in the TIAGo and IRB120.

Regarding the analysis of the joints' angle distribution, we studied whether the use of KGEs in the learning process might involve an improvement in the exploration ability and, in consequence, better exploitation and more efficient learning. We see that in the TIAGo setup, there is a significant drop in the standard deviation value for joint 4 and a big change in the joint 3 mean. Conversely, in the IRB120 scenario, we did not obtain great standard deviation variations, but differences in the distribution means, as we see in joint 2 and, especially, in joint 3. This absence of meaningful changes might be in concordance with the fact that the BM is already close to mastering the task, so the differences in the exploration are not that relevant. These outcomes show that under the setup without DR, the exploration capability might present an enhancement, helping the KGE agents to achieve better policies sooner than the BM agents and, hence, learn faster.

When we apply DR to the targets' color, we see that for the TIAGo and the IRB120, the agents with the complete KGE are three times faster than the other agents, reducing the learning time needed by almost \qty{60}{\percent}. Besides, their transitory regime is consistently better, as it happens in the experiments without DR. This might confirm that in a situation with an increase in the problem complexity resulting from varying features, adding contextual information helps in the learning process and the final performance. For the TIAGo, the full KGE agent performs \qty{16}{\percent} better than the partial KGE agent and the BM. The TIAGo's and IRB120's trendline analysis shows that the agent with the full KGE decreases its performance around the 30~M step and 50~M step, respectively, although later, they narrow the distance again with the other models. We argue that this might have happened due to a high learning rate and the fact that we do not use a learning rate decay mechanism.

For the IRB120, there is \qty{10}{\percent} improvement between the full KGE agent and the BM. With this robot, it is noticeable that there is not a significant improvement between the full KGE and partial KGE success rate as it is in the TIAGo. We suggest that the color information does not involve an accuracy increase because of the scenario colors. Since the robot is orange, the RGB filters corresponding to the green and, especially, the blue channels do not have the same relevance as the red ones. Hence, this unbalance in the visual learned features might implicate an exclusion of part of the color information encoded in the KGE. This does not happen with the TIAGo because its arm is white, so it should equally use the filters for the three RGB channels.

Finally, the outcomes obtained by analyzing the distribution of the joints' orientation, we noted a significant decrease in the standard deviation of the distribution for three of the TIAGo and IRB120 joints when KGEs are used. As in the previous set of experiments, this suggests that the agents can exploit the correct policies early thanks to the information encoded in the embeddings. This is consistent with the fact that these agents learn faster than the BM agents. This drop is larger and occurs in more joints than in the scenarios trained without DR, which might be because of the increase in the problem's complexity.

%%%%%%%%%%%%%%%%%%%%%%%%%%%%%%%%%%%%%%%%%%
\section{Limitations}
\label{sec:Limitations}
While our experiments have shown improved learning efficiency in predefined environments, the reliance on predefined knowledge graphs may limit adaptability in dynamic and unstructured environments, where the robot must perform tasks that cannot benefit from semantic conditioning.
The effectiveness of our approach depends on the completeness of the knowledge graph. If the graph lacks critical semantic relationships or contains incorrect relationships, the resulting embeddings may mislead the DRL agent, potentially degrading its performance. 
Our experiments were conducted in simulated environments to ensure controlled evaluations. However, since our observation is represented by an RGB image, transferring the learned policies to real-world robotic platforms would need an additional image-to-image translation method in order to efficiently bridge the simulation-to-reality gap.
%Addressing these limitations in future work will further enhance the applicability and robustness of DRL agents leveraging semantic knowledge for robotic control.

%%%%%%%%%%%%%%%%%%%%%%%%%%%%%%%%%%%%%%%%%%
\section{Conclusion}
\label{sec:Conclusion}
In this paper, we demonstrate how adding contextual information about the environment, encoded in embeddings, helps DRL agents learn faster and more efficiently. These descriptions of the environment are obtained through the use of embeddings obtained by the selection of semantically relevant nodes in a Knowledge Graph.
We have shown that adding these inputs to the DRL model does not significantly increase the computational time, and the impact on the model's architecture complexity is minimal. We propose an original architecture in which the embeddings are concatenated to the hidden activations of the FC layer that follows the representation learning blocks, thus complementing the visual information.

Overall, the results show an improvement in the agents' performance when KGEs are introduced. If the targets' colors are fixed, we noted a faster learning transitory regime, a decrease in the learning time, up to \qty{25}{\percent} in the TIAGo setup, and an increase in the agent's accuracy by \qty{12}{\percent} for both robots. If we apply DR to the targets' colors, the behavior of the transitory period remains better than the baseline, the decrease in the learning time rises up to \qty{60}{\percent} in both robots, and the agent's accuracy is increased up to a \qty{20}{\percent} for TIAGo and \qty{10}{\percent} for IRB120. Additionally, we investigated the effect of KGEs on the joints' angle distribution to determine if there is an improvement in the agents' ability to exploit sooner the correct policy between the full KGE agent and the baseline. 

For the experiments without DR, there are changes in the mean and standard deviation for some joints, but these are higher, and more joints are affected in the scenarios with DR. This suggests that, first, the use of KGEs enhances the agents' learning process and therefore, the learning time is reduced, the exploitation of the best policies happens earlier, and higher accuracies are achieved; and second, that providing the embedded information in a more complex problem leads to a more significant improvement than in simpler scenarios.

In conclusion, we have validated the proposed architecture that integrates contextual information about the environment, given as KGEs. We have proven that if DRL agents can leverage semantic knowledge, there is an improvement in their learning speed and success rate without significant increases in computational and architectural complexity. 

Beyond the proposed application, the presented methodology shows promising capabilities in any scenario where DRL agents can leverage semantic knowledge sources. 
We leave for future work the analysis of how the agents' performance is affected depending on where the KGE input is placed, as in our current approach the placement has been selected after the state representation layers, but different integration points within the network may influence how effectively the semantic information is combined with the learned visual features. Besides, the next research steps also include the sim-to-real transfer for both setups, looking for a zero-shot or few-shot transfer where we can reuse, completely or partially, the knowledge learned in the virtual scenarios. To do so, we should explore domain adaptation techniques that minimize or potentially eliminate the need for real-world experience.

%%%%%%%%%%%%%%%%%%%%%%%%%%%%%%%%%%%%%%%%%%
\vspace{6pt} 

%%%%%%%%%%%%%%%%%%%%%%%%%%%%%%%%%%%%%%%%%%
\section*{Author contributions}
Conceptualization, L.G-L., V.S., J.B., A.L-L. and D.N.; methodology, L.G-L, V.S., J.B., A.L-L. and D.N.; software, L.G-L and V.S.; validation, L.G-L and V.S.; formal analysis, L.G-L and V.S.; investigation, L.G-L and V.S.; resources, L.G-L and V.S.; data curation, L.G-L and V.S.; writing---original draft preparation, L.G-L and V.S.; writing---review and editing, L.G-L, V.S., J.B., A.L-L. and D.N.; visualization, L.G-L and V.S.; supervision, J.B., A.L-L. and D.N. All authors have read and agreed to the published version of the manuscript.

\section*{Declarations}
\begin{itemize}
	\item Funding: This research did not receive any specific grant from funding agencies in the public, commercial, or not-for-profit sectors.
	\item Conflict of interest/Competing interests: All authors certify that they have no affiliations with or involvement in any organization or entity with any financial interest or non-financial interest in the subject matter or materials discussed in this manuscript.
	\item Availability of data and materials: Researchers or interested parties are welcome to contact the corresponding author L.G-L. for further explanation.
	\item Code availability: Researchers or interested parties are welcome to contact the corresponding author L.G-L. for further explanation.
\end{itemize} 

\bibliographystyle{elsarticle-harv} 
\bibliography{bibliography}

\end{document}